\documentclass[letterpaper, 10 pt, conference]{ieeeconf}  % Comment this line out if you need a4paper

\IEEEoverridecommandlockouts                              % This command is only needed if 
\usepackage{graphics} % for pdf, bitmapped graphics files
\usepackage{epsfig} % for postscript graphics files
\usepackage{mathptmx} % assumes new font selection scheme installed
\usepackage{times} % assumes new font selection scheme installed
\usepackage{amsmath} % assumes amsmath package installed
\usepackage{amssymb}  % assumes amsmath package installed
\usepackage{subfigure}
\usepackage{multirow}
\usepackage{graphicx, epstopdf}
\usepackage{xspace,epsfig,url}
\usepackage[noadjust]{cite}
\usepackage{enumerate}
\usepackage{float}
\usepackage{epsf}
\usepackage{psfrag}
\usepackage{verbatim}
\usepackage[all]{xy}
\usepackage{color}
\usepackage[table]{xcolor}
\usepackage{bm}
\usepackage{cases}
\usepackage{hyperref}
\usepackage[ruled,vlined,linesnumbered]{algorithm2e}

\title{\LARGE \bf
Human Interface for Teleoperated Object Manipulation\\ with a Soft Growing Robot
}

\author{Fabio Stroppa, Ming Luo, Kyle Yoshida, Margaret M. Coad, \\ Laura H. Blumenschein, and Allison M. Okamura% <-this % stops a space
\thanks{Toyota Research Institute (TRI)  provided funds to assist the authors with their research but this article solely reflects the opinions and conclusions of its authors and not TRI or any other Toyota entity.}% <-this % stops a space
\thanks{The authors are with the Mechanical Engineering Department, Stanford University, Stanford, CA 94305, USA
        {\tt\small fstroppa@stanford.edu}}%
}

\begin{document}

\maketitle
\thispagestyle{empty}
\pagestyle{empty}

%%%%%%%%%%%%%%%%%%%%%%%%%%%%%%%%%%%%%%%%%%%%%%%%%%%%%%%%%%%%%%%%%%%%%%%%%%%%%%%%
\begin{abstract}

Soft growing robots are proposed for use in applications such as complex manipulation tasks or navigation in disaster scenarios. Safe interaction and ease of production promote the usage of this technology, but soft robots can be challenging to teleoperate due to their unique degrees of freedom. 
In this paper, we propose a human-centered interface that allows users to teleoperate a soft growing robot for manipulation tasks using arm movements. 
A study was conducted to assess the intuitiveness of the interface and the performance of our soft robot, involving a pick-and-place manipulation task.
The results show that users completed the task with a success rate of $97\%$, achieving placement errors below $2$~cm on average. These results demonstrate that our body-movement-based interface is an effective method for control of a soft growing robot manipulator.

\end{abstract}

%%%%%%%%%%%%%%%%%%%%%%%%%%%%%%%%%%%%%%%%%%%%%%%%%%%%%%%%%%%%%%%%%%%%%%%%%%%%%%%%
\section{INTRODUCTION}

% introduction to soft robots
Soft and continuum robots have useful features that are advantageous in applications requiring delicate interaction, e.g.\ object manipulation \cite{trivedi2008soft,calisti2011octopus, coevoet2019soft, brown2010universal,ilievski2011soft,cianchetti2014soft}, or adaptation to unknown environments, e.g.\ navigation and exploration \cite{coad2019vine,wooten2018exploration}. A subset of soft and continuum robots have an additional feature that makes operation in confined environments easier: the ability to extend or grow as an additional degree of freedom \cite{hawkes2017soft, coad2019vine,wooten2018exploration,gilbert2016concentric}. By extending and shortening in length, these systems can move their tip through cluttered environments without being restricted by body parts that may collide with obstacles, such as the ``elbows" on a typical rigid serial-chain robot arm. For this reason, growth can be especially beneficial in manipulation tasks.
%Continuum robots, or soft-growing robots, are currently being used in a large number of scenarios where mechanical robots would be less efficient \cite{walker2013continuous}. 
%Applications may vary from navigation \cite{webster2010design, murphy2016disaster, liu2013state, coad2019vine} to object manipulation \cite{calisti2011octopus, coevoet2019soft}: in all these examples, the flexible structure of the soft robots allows them to overcome several mechanical limitations \cite{yim2003modular}. Furthermore, soft robots offer the potential for safe physical interactions with humans, while safety in rigid robots is limited by inherent mechanical properties \todo{[FIND REF]}. 
%When used to manipulate objects, soft robots can achieve good performance due to their growth by eversion property: this allows them to move in a wide workspace without being constrained by mechanical joints, especially when working in constrained and cluttered environments \cite{neppalli2007octarm, mehling2006minimally}.
%While the growth degree of freedom has benefits in cluttered environments, in the case of designing human-in-the-loop teleoperation for these extending soft robots, designing an intuitive teleoperation interface that allows the user to leverage those benefits is challenging.

While the growth degree of freedom has benefits in cluttered environments, designing control to leverage those benefits is challenging. In general, there do not exist well-defined kinematic models for soft robots, so often control of soft robots happens in joint space instead of task space \cite{rus2015design}. Even when approximate kinematic models exist, the output shape or behavior of the robot can be difficult to measure, and therefore hard to close a loop around. Thus, one strategy to control soft robotic systems is to use the human to close the loop on position and account for errors caused by inaccurate models and lack of sensing and closed-loop control. 
However, dissimilarity between the degrees of freedom of the robot and the human makes it difficult to find appropriate control interfaces.

% problem of the interface
Studies have used devices such as 3D mice \cite{fellmann2015evaluation}, joysticks and gamepads for gaming \cite{csencsits2005user, grissom2006design, fellmann2015evaluation}, haptic interfaces \cite{fellmann2015evaluation , majewicz2013cartesian}, rigid-link manipulators \cite{frazelle2016teleoperation}, and even flexible joysticks 
specially designed for soft robots \cite{el2018development}. In particular, the work of El-Hussieny et al. \cite{el2018development} was specifically designed for soft growing robots and proved to be intuitive and easy to use. 
%; but due to the kinematic dissimilarity between these interfaces and a soft robot, the performance can be negatively affected.
%One of most intuitive and performing interfaces is the flexible joystick proposed by El-Hussieny et al. \cite{el2018development}, which maps the human bending commands into movements and provides an implicit shape sensing capability for a distal portion of the soft robot. 
However, all these interfaces rely on physical devices, which may not be the most intuitive way for humans to control (and learn to control) the robot.  %[I softened this statement because the user a body-interface device also has to learn commands]
Here we remove the physical interface and use the human body to control the robot.

\begin{figure}[t!]
  \centering
	{\includegraphics[width=\linewidth]{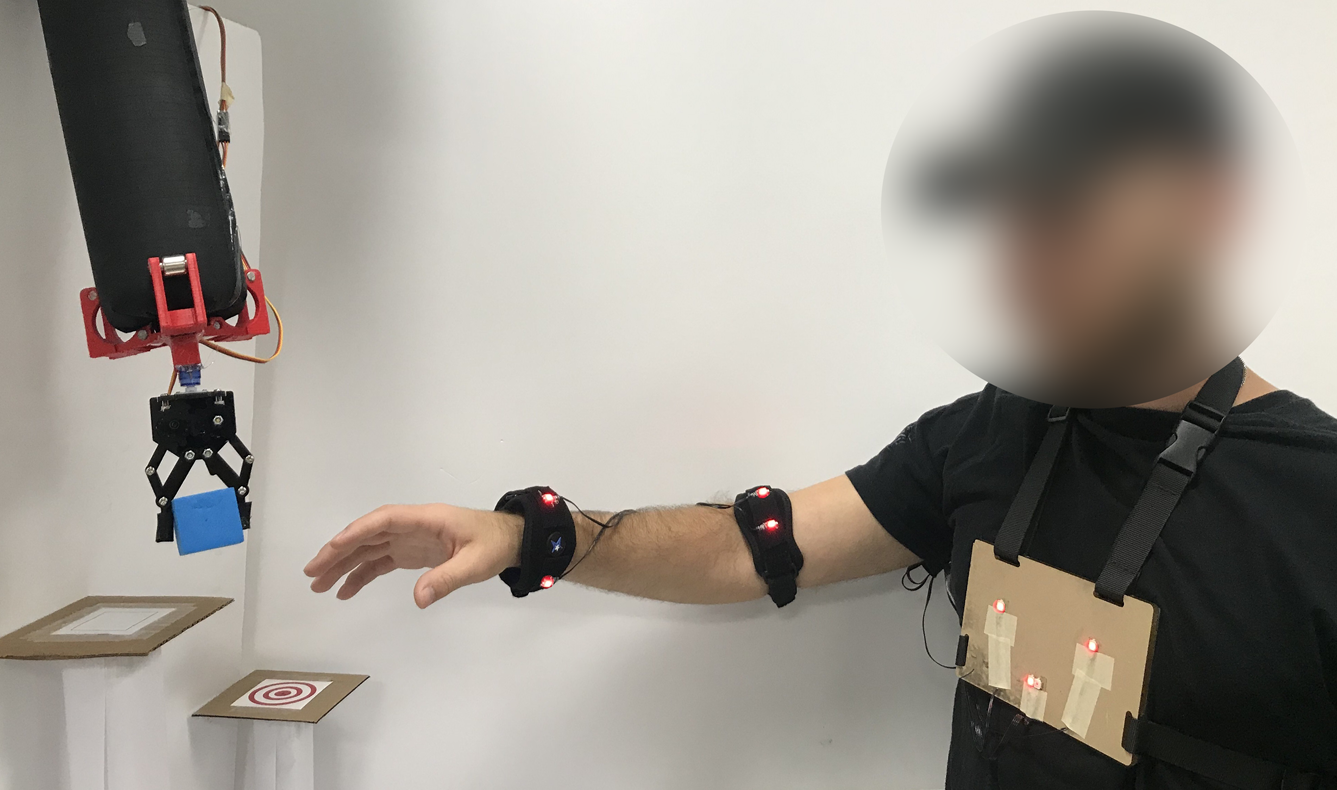}}
    \caption{An operator controls the soft growing robot with the gesture-based Body Interface. Here the operator is shown physically near the robot, which is safe due to the robot's low inertia and soft exterior, while in our experimental study the operators controlled the robot from a slightly farther distance.}
    \label{fig:operator_and_robot}
\end{figure}

% proposal
In this work, we propose an interface that allows human operators to control the robot simply by using their arm. The gestures of the operator, tracked by a motion capture system, are mapped to the kinematics of the robot for an easy and intuitive teleoperation. This interface, called the ``Body Interface'' (Fig.~\ref{fig:operator_and_robot}), was used in an experimental study to assess its effectiveness in the control of a soft growing robot in a teleoperated manipulation task. Twelve participants were able to successfully teleoperate the robot to reach, grasp, and move objects in the workspace.
%\todo{Furthermore, to the best of our knowledge, this is the first study presented in literature that involves real soft robots performing manipulation tasks.}

% structure of the work
The rest of the paper is organized as follows: Sec.~\ref{sec:bodyinterface} discusses the interface, Sec.~\ref{sec:setup} describes the design and control of the soft growing robot, Sec.~\ref{sec:experiment} discusses the experiment setup and results, and, finally, Sec.~\ref{sec:conclusion} summarizes the work and presents possible future research.

\begin{figure*}[t!]
\vspace{-0.5cm}
\centering
    \subfigure[\protect\url{}\label{fig:markers}]%
	{\includegraphics[height=6.2cm]{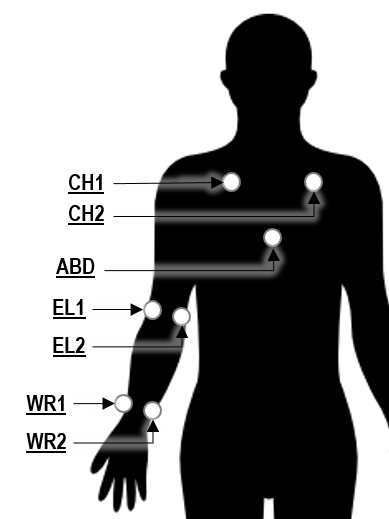}}\qquad
	\subfigure[\protect\url{}\label{fig:commands}]%
	{\includegraphics[height=6.2cm]{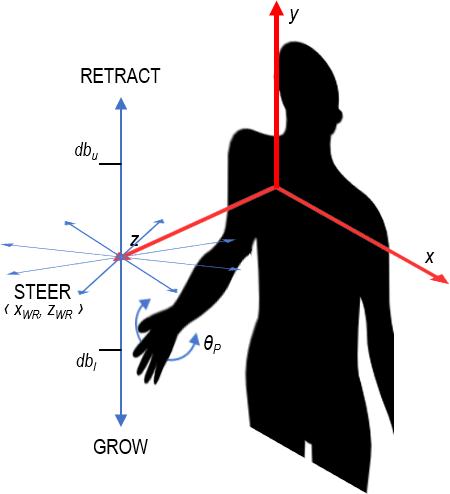}}\qquad
	\subfigure[\protect\url{}\label{fig:ref_systems}]%
	{\includegraphics[height=6.2cm]{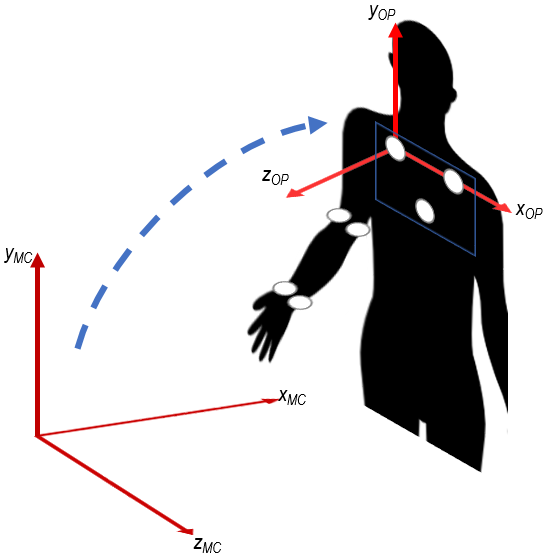}}
	\caption{(a) Motion capture marker layout on the operator's upper body. (b) Commands used to control the robot based on direction of movement. (c) The original reference system is transformed to be aligned with the plane where the markers \textit{CH1}, \textit{CH2}, and \textit{ABD} lie, with origin at \textit{CH1}.}
	\label{fig:body_interface}
	\vspace{-0.5cm}
\end{figure*}

\section{BODY INTERFACE}
\label{sec:bodyinterface}

The interface for teleoperating the robot, the Body Interface, is based on a Motion Capture Tracking system and a Gesture Interpreter Tool. The interface tracks the operator's gestures, maps them to the kinematics of the robot, and sends the commands.

\subsection{Motion Capture Tracking}

We used the \textit{PhaseSpace Impulse X2E} (phasespace.com) system to track the operator's movements. This accurate optical tracking mechanism was used in order to test the effectiveness of the interface while avoiding the performance limitations of other types of sensors. In practice, other tracking systems such as inertial measurement units (IMUs) could be employed.

Our motion capture setup includes six lightweight-linear-detector cameras monitoring seven active LED markers placed on the forearm and the chest of the operator. 
The gestures are tracked in real time at $270~Hz$. %, exploiting the PhaseSpace system technology that eliminates marker-swapping for cleaner data and high precision.
As shown in Fig.~\ref{fig:markers}, the Body Interface exploits four markers on the operator's forearm for gesture recognition (two on the elbow \textit{EL1} and \textit{EL2}, two on the wrist \textit{WR1} and \textit{WR2}), and three on the operator's chest to create a body centered reference system (\textit{CH1}, \textit{CH2}, and \textit{ABD}).

% \begin{figure}[t!]
%   \centering
% 	{\includegraphics[height=7cm]{figures/markers.png}}
%     \caption{Marker layout.}
%     \label{fig:markers}
% \end{figure}

\subsection{Gesture Interpreter Tool}

The operator's gestures are mapped to the kinematics of the robot through our custom Gesture Interpreter Tool (GIT). The GIT recognizes three types of commands: grow/retract, steer left/right/backwards/forwards, and rotate the end effector. One command of each type can be given simultaneously. The communication with the robot's microcontroller is realized via a serial port at $66~Hz$.

The specific mapping between the gestures and commands can be customized based on the application. Fig.~\ref{fig:commands} shows one proposed mapping, used to control a soft robot hanging from the ceiling and growing in the direction of gravity.
Moving the forearm above and below the operator's transverse plane (forearm flexion/extension) will make the robot retract and grow, respectively; whereas all the movements parallel to the transverse plane are mapped as steering movements (forearm back and forth and medial/lateral rotation, respectively backwards/forwards and left/right); finally, pronosupination defines the end effector rotation. 

\subsubsection{Calibration and Command Mapping}

The location of the wrist and elbow define the sent commands. Since the interface is based on body movements, the system needs an initial calibration to account for the operator's reach workspace.

The steering commands are mapped to the $x$ and $z$ coordinates of the wrist. The wrist location \textit{WR} (the centroid of \textit{WR1} and \textit{WR2}) is projected to the operator's transverse plane to give the coordinates $\langle x_{WR}, z_{WR}\rangle$. 
During the calibration, the system stores the limits of the operator's reach in the four directions (left/right/backwards/forwards), which will then correspond to the limits of the robot's workspace.

% \begin{figure}[t!]
% \centering
% 	\subfigure[\protect\url{}\label{fig:commands}]%
% 	{\includegraphics[height=7cm]{figures/commands_p2.png}}\qquad
% 	\subfigure[\protect\url{}\label{fig:learning_2}]%
% 	{\includegraphics[height=7cm]{figures/ref_systems.png}}
% 	\caption{\textit{(a)} List of commands used to control the robot. \textit{(b)} The original reference system is transformed to be aligned with the plane where \textit{CH1}, \textit{CH2}, and \textit{ABD} lie, with origin in \textit{CH1}.}
% 	\label{fig:learning}
% \end{figure}

The command of growth/retraction is triggered when the operator's hand exceeds a certain threshold of $y_{WR}$. During calibration, the operator defines a deadband ($[db_l, db_u]$) along the $y$ axis: if the $y$ coordinate of \textit{WR} falls within this region, the robot keeps its length fixed; otherwise, it grows or retracts at a fixed speed, based on the position of the operator's hand. In this case, the calibration of the deadband is defined by half of the operator's reachable limits, to allow the operator to easily steer and change length simultaneously.

Finally, the angle $\theta_P$ defines the rotation of the end effector. This is the angle between the two segments \textit{WR1}$-$\textit{WR2} and \textit{EL1}$-$\textit{EL2} when their projection lies on the operator's coronal plane. During calibration, the offset between $\theta_P$ and the starting orientation of the end effector is stored to assure the operator's comfort during the teleoperation.

\subsubsection{Reference System Alignment}
In order to properly retrieve the data, the GIT needs to define a body centered reference system.
The three chest markers allow the operator to be aligned to the motion capture reference system, resulting in an interface that is independent of the operator's pose in space. 
As shown in Fig.~\ref{fig:ref_systems}, the frame defined by the calibration of the Motion Capture system $\langle x_{MC}$, $y_{MC}$, $z_{MC} \rangle$ is transformed into the reference system of the operator $\langle x_{OP}$, $y_{OP}$, $z_{OP} \rangle$, such that the coordinates of the markers are expressed with reference to the latter.
In particular, \textit{CH1}, \textit{CH2}, and \textit{ABD} define a plane, which the GIT transforms to be lying on the $x_{OP}$-$y_{OP}$ plane, with \textit{CH1} placed at the origin of the new reference system. 
%By applying this transformation to all the markers, the reference system is changed.
The operator can therefore control the robot in whatever body pose is most comfortable.

\section{SOFT-GROWING ROBOT}
\label{sec:setup}

We built a soft growing robot specifically for manipulation tasks. This section describes its design and control strategies.

\begin{figure}[t!]
\centering
    \subfigure[\protect\url{}\label{fig:taskSchematic_1}]%
	{\includegraphics[height=5.91cm]{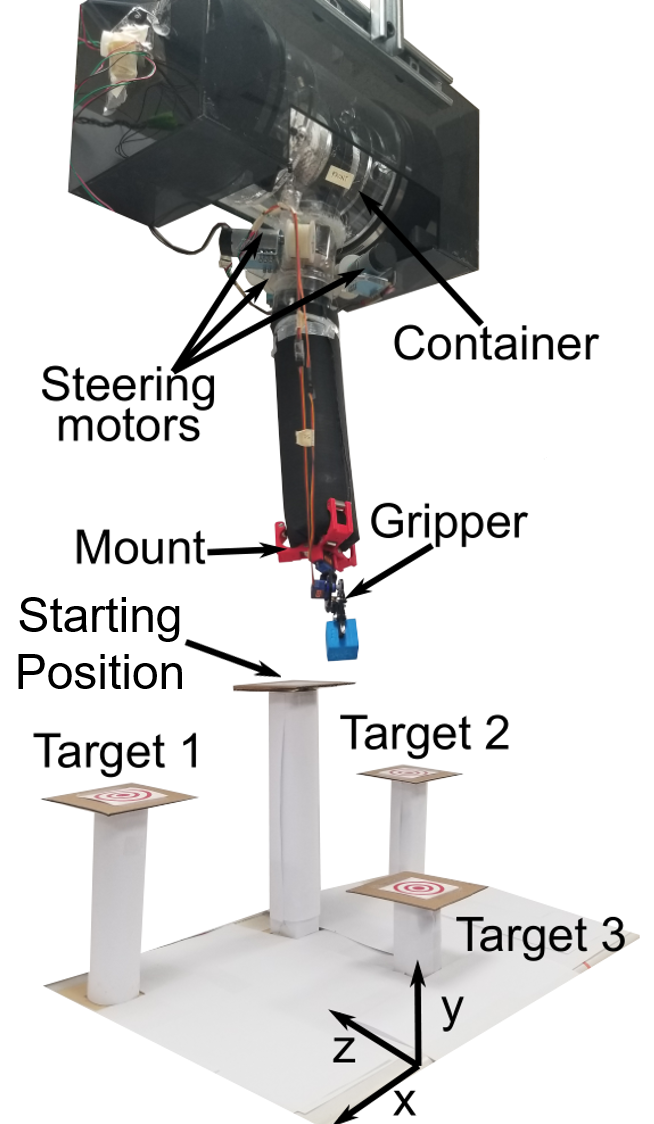}}\qquad
	\subfigure[\protect\url{}\label{fig:taskSchematic_1}]%
	{\includegraphics[height=5.91cm]{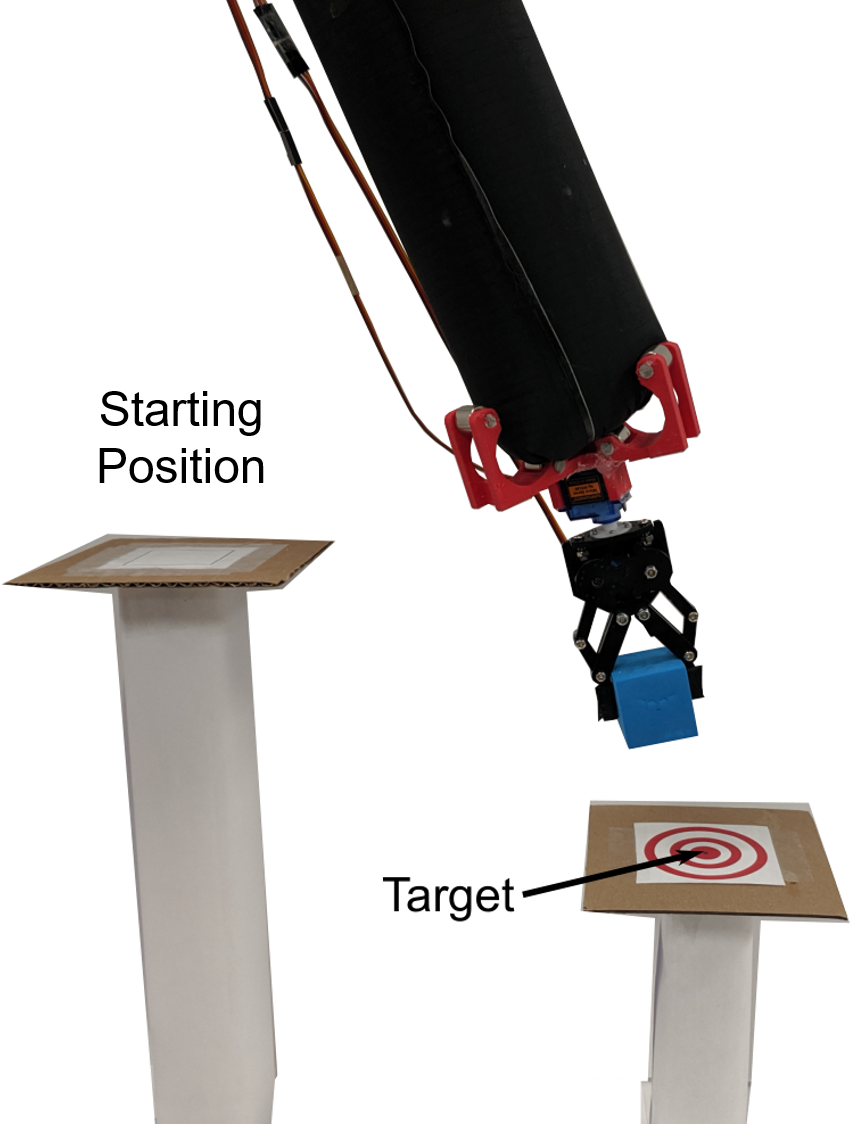}}
	\caption{(a) The soft growing robot with its components, and the proposed task scenario with the orientation of the operator's reference system. (b) The soft robot during the manipulation task, moving a block to a specific target.}
	\label{fig:soft_robot}
	\vspace{-0.5cm}
\end{figure}

\subsection{Design}

The soft growing manipulator can grow, retract, and steer in three dimensions while carrying a payload, as shown in Fig.~\ref{fig:soft_robot}. 
The device retracts into a portable, sealed container which can be easily mounted anywhere.
The soft growing manipulator everts, adding new material at the tip when pressurized, and the DC motor inside the container pulls at the tip, inverting the material for retraction. The fabrication of the soft growing manipulator and the design of the manipulator's container are similar to the robot described in our previous work \cite{coad2019vine}, with the addition of two new components: (i) a cable-driven steering system, and (ii) a wireless gripper mount. The steering system consists of three evenly spaced cables, driven by three gearmotors (\textit{Pololu 131:1 37DX73LM}) at the container outlet to provide 3-degree of freedom motion. The ends of the cables are fixed to the proximal end of the manipulator for steering. The gripper is driven by two servo motors. One controls the rotation and the other controls grasping. The gripper (\textit{Standard Gripper Kit-Rollpaw, SunFounder}) connects to and moves along the tip of the robot with a magnetic attachment similar to the one presented in \cite{luong2019eversion}, allowing for the completion of grasping and manipulation tasks. 
The robot is made of a heat-sealable thermoplastic polyurethane fabric sheet and can grow to up to $1.5$~m, with a diameter of $10$~cm.

\subsection{Control}

With reference to the parameters described in Sec. \ref{sec:bodyinterface}, the Body Interface controls the following robot parameters:

\begin{itemize}
    \item the end effector position (in meters), given by $\langle x_{WR}, z_{WR} \rangle$, these are the two coordinates of the tip of the robot given a certain length of the body\footnote{The coordinate along the direction of growth may vary when the length of the robot is fixed, as a result of steering.};
    
    \item the orientation of the gripper (in radians), given by the angle $\theta_P$; and
    
    \item the direction of length change, either growing, retracting, or static, given by $y_{WR}$ relative to the growth deadband $[db_l, db_u]$. When the deadband is exceeded, the robot is commanded to grow or retract at a constant rate (in radians per second)\footnote{Since growth is driven by internal pressure in addition to the container motor, the actual robot growth is not constant but is upper bounded by the commanded motor speed.}. %as shown in (\ref{eq:BI_s}).
\end{itemize}
%\vspace{1cm}
%\begin{align}\label{eq:BI_s}
%	\bigg \{
%	\begin{array}{ll}
%		db_u < y_{WR} \qquad \qquad \quad \rightarrow \quad -k_{sp} \\
%		y_{WR} < db_l \qquad \qquad \quad \rightarrow \qquad k_{sp} \\
%		db_l \leq y_{WR} \leq db_u \qquad \ \rightarrow \qquad 0 \\
%	\end{array}
%\end{align}
%
%&The term $k_{sp}$ in (\ref{eq:BI_s}) is the constant of speed, which was fixed at $100~tps$ such that the robot is commanded to grow a constant rate\footnote{\todo{The actual robot growth is not constant due to the uncertainty present in soft robot systems.}}.

Because the robot tip does not have tracking sensors, the controller relies on the human operator to close the loop and achieve the desired end effector position.
More details about the mapping and control strategies can be found in our previous work \cite{coad2019vine}, which is based on the constant curvature model of continuum robots \cite{webster2010design}.

%Note that, for the current setup, the grasping command is given by pressing an external button: as Sec. \ref{sec:conclusion} will illustrate, future work will look at adding a holdable device for haptic guidance \cite{walker2019holdable} that the operator will use to both receive haptic feedback about the robot state and the environment, and to send the grasping commands. 
For the experiment described in Section~\ref{sec:experiment}, the operator opens and closes the gripper with a verbal command to the investigator.

\section{EXPERIMENTAL STUDY}
\label{sec:experiment}

The Body Interface was tested on a pick-and-place task to evaluate its usability in terms of accuracy, timing, and workload. 

\subsection{Participants}

Twelve participants took part in the experiment (seven males, $26\pm3$ yrs old; and five females, $22\pm3$ yrs old). All participants were right handed and had no known impairment affecting their upper limb. The experimental protocol was approved by the Stanford University Institutional Review Board, and written informed consent was obtained from each participant.

\subsection{Task and Scenario}

%no abbrev. at the beginning of a paragraph, so I changed Fig. to Figure here.
Figures~\ref{fig:operator_and_robot} and ~\ref{fig:soft_robot} show the scenario of the experiment. The participants were asked to pick up a block placed on a starting pillar underneath the robot's base and then move the block onto a designated target. There were three targets placed over three different pillars, and all the participants repeated the task five times for each target, for a total of fifteen repetitions. The experiment was designed such that the participants were randomly and equally divided to explore all the six combinations of target ordering.

The setup details, including elements, size, and layout, are as follows:
\begin{itemize}
    \item the \textit{Block}, size $3.4 \times 3.4$~cm, placed at the center of the workspace on a pillar $30$~cm tall;
    \item \textit{Target 1}, placed $25$~cm to the left of the block, on a pillar $20$~cm tall;
    \item \textit{Target 2}, placed $25$~cm to the right of the block, on a pillar $17$~cm tall;
    \item \textit{Target 3}, placed $25$~cm in front of the block, on a pillar $9$~cm tall; and
    \item the \textit{Robot's Base}, placed over the block, a distance of $1$~m from the ground and $70$~cm from the block.
\end{itemize}
The block starting location and the targets were placed in the center of support surfaces with an area of $12\times12$~cm, and the robot started each trial from an initial length of $50$~cm measured from gripper to container. 
The participants were asked to face the robot as shown by the reference system in Fig.~\ref{fig:soft_robot}. Note that this was not a necessary constraint, as the GIT fixes the reference system based on the operator's position, but it was useful to normalize the position of the participants among all the trials and assure consistency in the results.

All the participants performed the experiment after a five-minute training phase, in which they were familiarized with the robot and the interface. They were instructed to move the robot, learn how fast the commands could be performed, explore the workspace (including testing the response of small and large hand movements), and grasp the block. During the training, the investigator illustrated strategies to get a good grasp on the block and retract without buckling the robot body. 

During the experiment, a trial was considered a \textit{failure} if the block fell to the table surface, which might be due to bad grasping, or hitting the pillar or the block and causing the block to move. This most often occurred after overshooting on the growth length. 
Participants were asked to repeat any failed trials, such that each of them performed a total of fifteen good trials.

After the training session, the participants started the real task, changing the target every five trials based on their designed ordering. In particular, a single trial was composed of two phases:
\begin{itemize}
    \item \textit{Grasping} phase, where the operator is asked to reach the block and grasp it; and
    \item \textit{Placing} phase, where the operator is asked to move the block from the starting position and place it in the designed target.
\end{itemize}
These phases were executed sequentially without a break in the participant's control of the robot, and both of them involved activities such as growing towards the targets, orienting the gripper for a proper grasp, avoiding the pillars, and retracting when needed (especially after grasping the cube to avoid dragging it on the support surface).
After each trial, the robot was automatically reset to its starting position and the block was manually replaced on the initial pillar by the investigator.

\subsection{Evaluation Metrics}

We used the following metrics to evaluate the task performance:

\begin{itemize}
    \item \textit{Target Placement Error (TPE)}: the distance, measured in centimeters, between the center of block and the target once the task is finished, representing the accuracy of the placement.
    \item \textit{Task Completion Time (TCT)}: the time required to complete a trial, measured in seconds. We broke this parameter into:
    (i) the overall time of the trial, from start to end;
    (ii) the time of each phase within a single trial; 
    and (iii) the time spent performing the actual grasp or placement, excluding the time spent in reaching either the block or the target.
    % \item \textit{Path Length (PL)}: the summation of the distances between two consecutive sampled points of the participant hand's movements, measured in millimeters.
    % \item \todo{\textit{Number of Grasps (NGP)}: the number of grasp attempts performed during the Grasping phase, including the successful one.}
    % \item \todo{\textit{Number of Growths (NGW)}: the number of growths performed during the task.}
    % \item \todo{\textit{Number of Retractions (NR)}: the number of retractions performed during the task.}
    \item \textit{Failure Rate (FR)}: the number of trials in which the block did not reach the target, which were then repeated.
    \item \textit{Standard NASA Task Load Index (NASA-TLX)}: a subjective standard assessment rating perceived workload while performing a certain task \cite{hart1988development}; the participants were asked questions about mental load, temporal load, effort, and frustration scale for each session, and the weighted average of these was used to calculate the overall workload.
\end{itemize}

\subsection{Results and Discussion}

%Error histogram 

\begin{figure}[b!]
%\vspace{-0.5cm}
  \centering
	{\includegraphics[width=\linewidth]{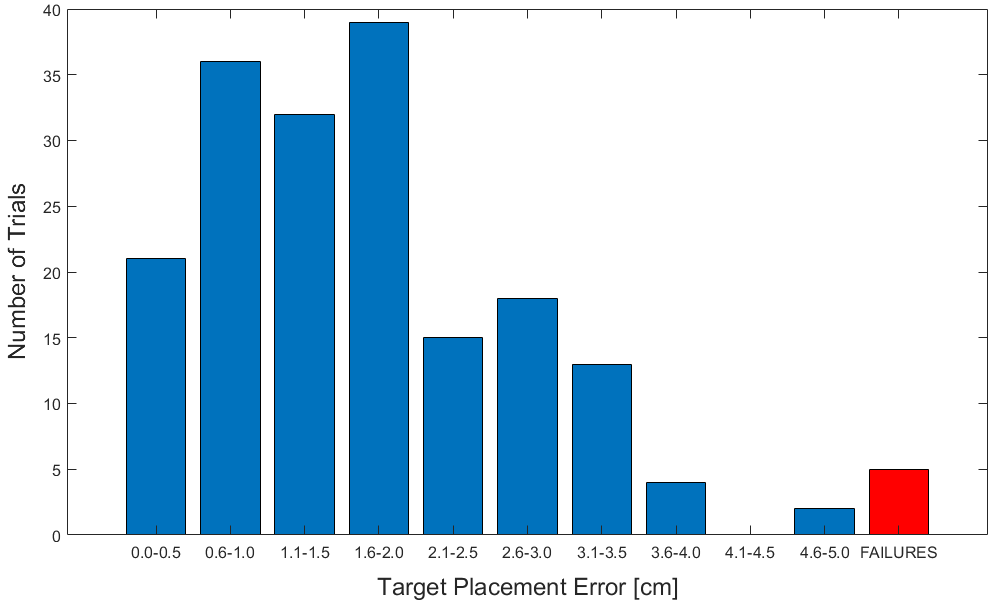}}
    \caption{Histogram of the Target Placement Error (TPE) over the all 180 trials (15 trials each for 12 participants), including all the successful block placements of the participants (in blue), compared to the number of failures (in red).}
    \label{fig:errorHistogram}
\end{figure}

\begin{figure*}[t!]
\centering
	\subfigure[\protect\url{}\label{fig:learning_1}]%
	{\includegraphics[width=8cm]{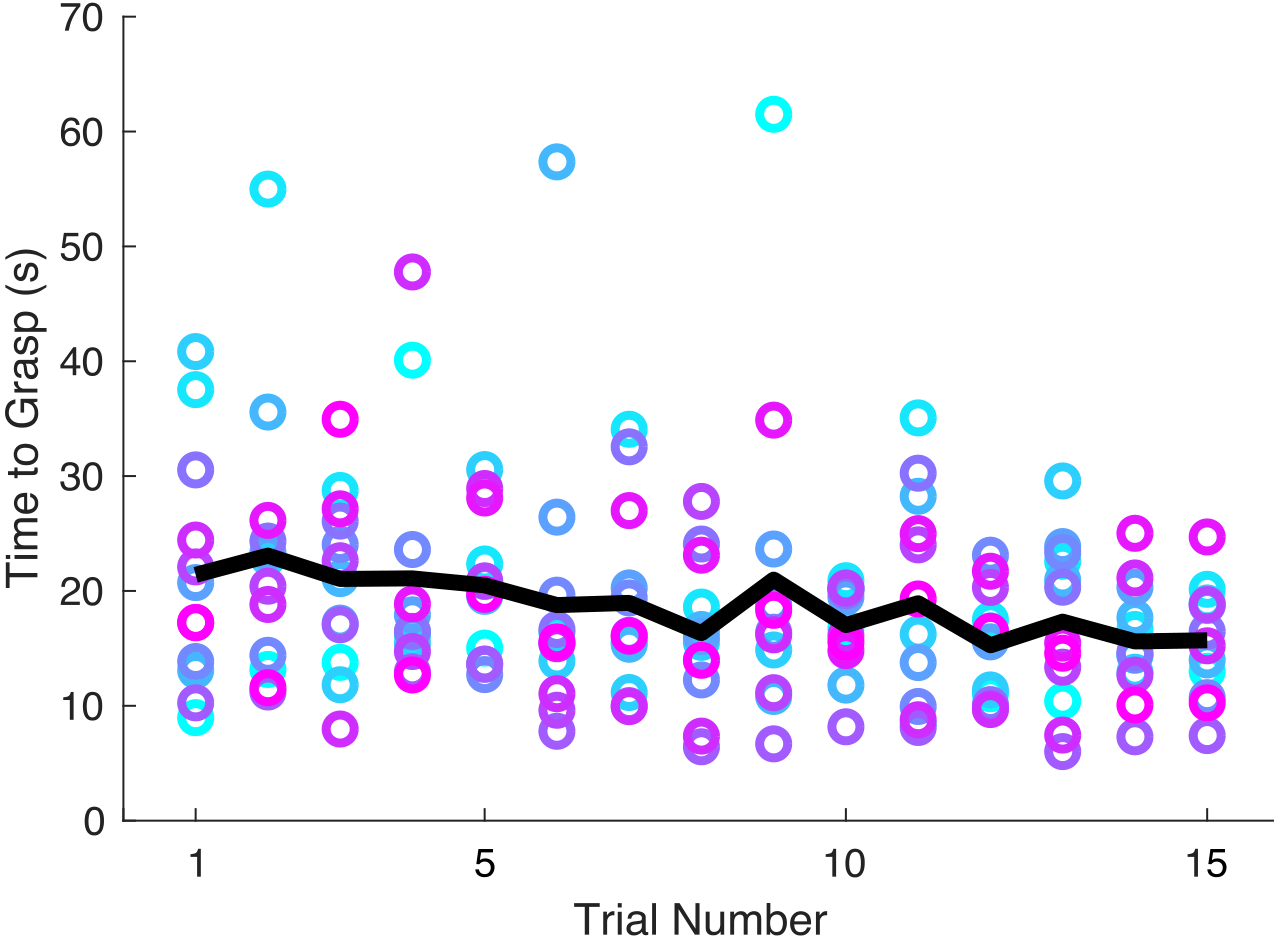}}\qquad
	\subfigure[\protect\url{}\label{fig:learning_2}]%
	{\includegraphics[width=8cm]{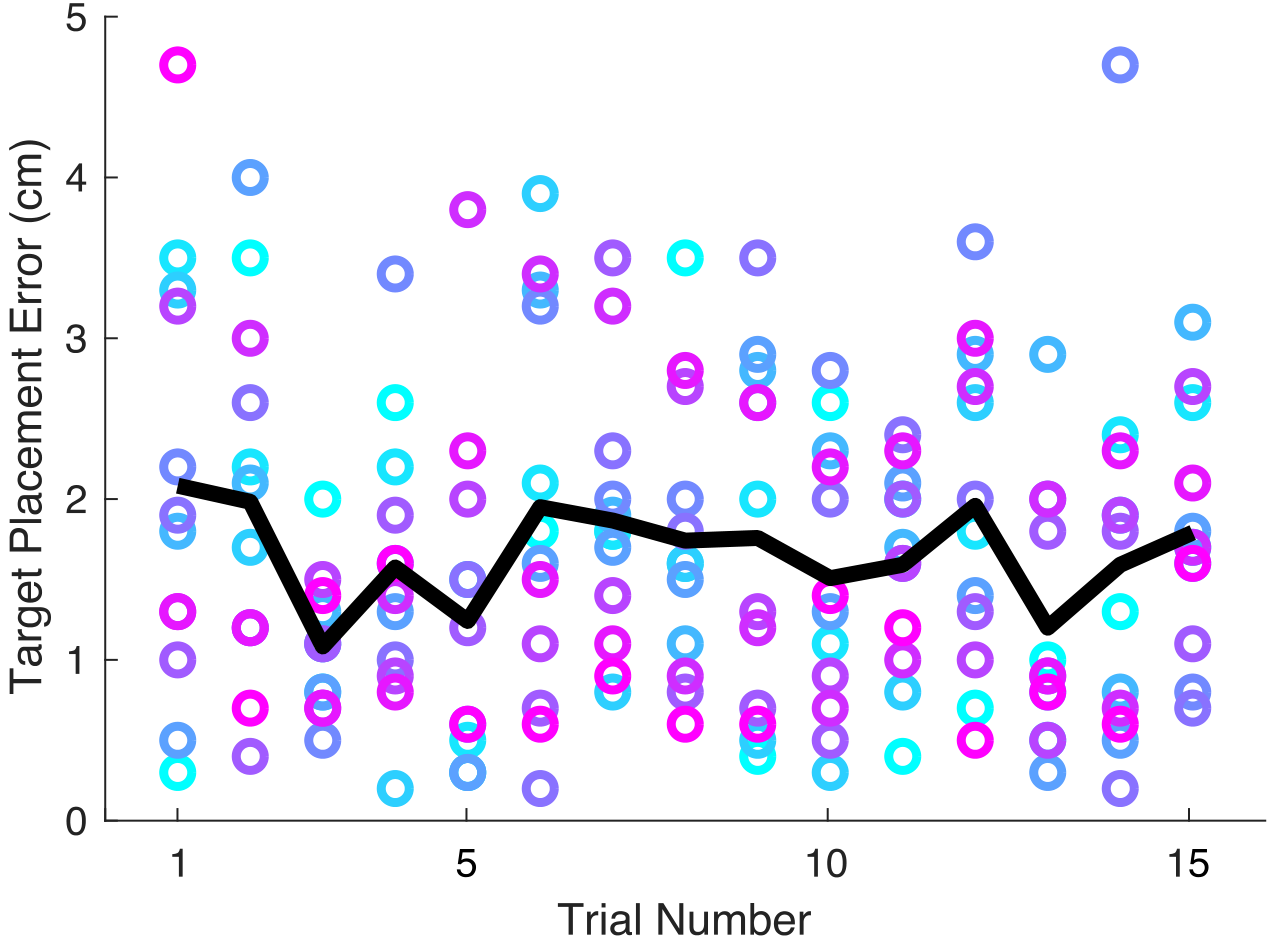}}
	\caption{Performance of the manipulation task throughout the experimental study. Dots show individual participants' performance (color corresponds to participant). Average performance across participants (black line) is consistent over the experiment. (a) Grasping phase performance as measured by time to successfully grasp the object. (b) Placing phase performance as measured by Target Placement Error (TPE).}
	\label{fig:learning}
	\vspace{-0.5cm}
\end{figure*}

\begin{figure*}[h!]
\centering
	\subfigure[\protect\url{}\label{fig:movT1}]%
	{\includegraphics[height=5cm]{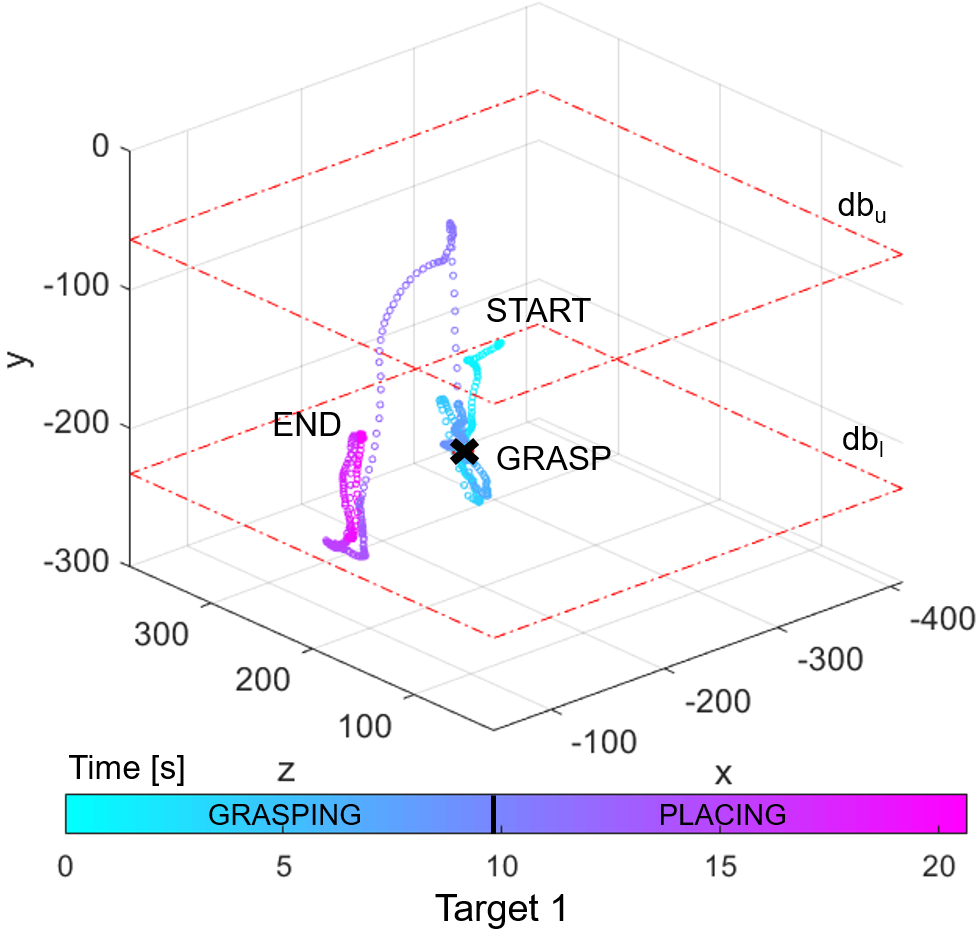}}\qquad
	\subfigure[\protect\url{}\label{fig:movT2}]%
	{\includegraphics[height=5cm]{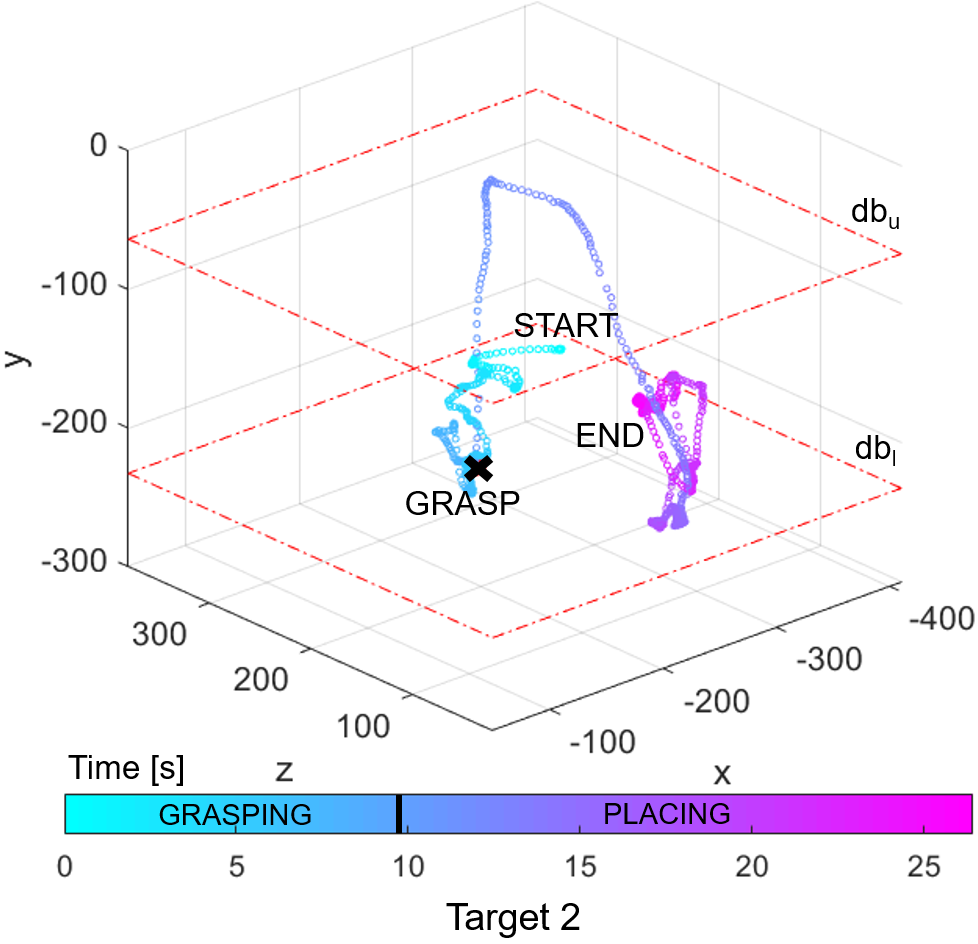}}\qquad
	\subfigure[\protect\url{}\label{fig:movT3}]%
	{\includegraphics[height=5cm]{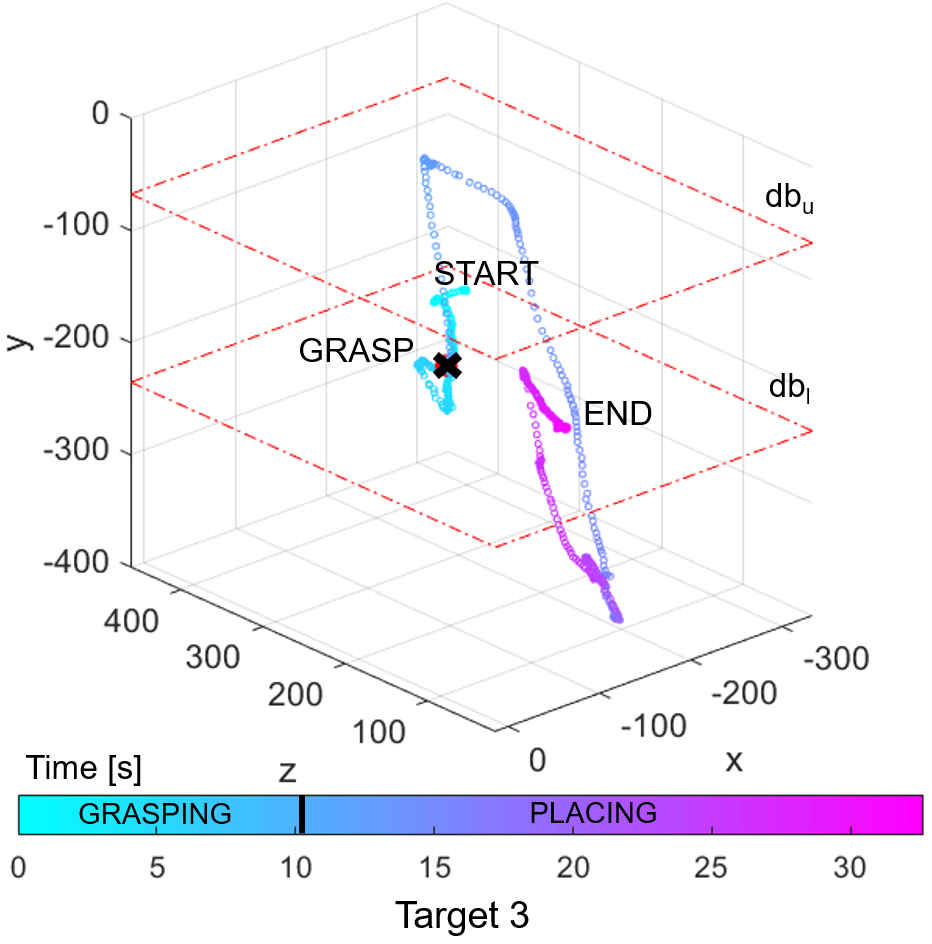}}
	\caption{Paths followed by the operator's hand during object manipulation trials, depicting the participant who completed the task in the shortest time for each target. Each plot indicates the starting and ending point, as well as the moment when the grasp was performed (the black $\times$). The planes outlined in red dashed lines represent the deadband $[db_l, db_u]$, indicating where the grow/retraction commands were triggered.}
	\label{fig:executionPath}
	\vspace{-0.5cm}
\end{figure*}

\begin{figure}[t!]
\centering
	\subfigure[\protect\url{}\label{fig:tct_total}]%
	{\includegraphics[width=8cm]{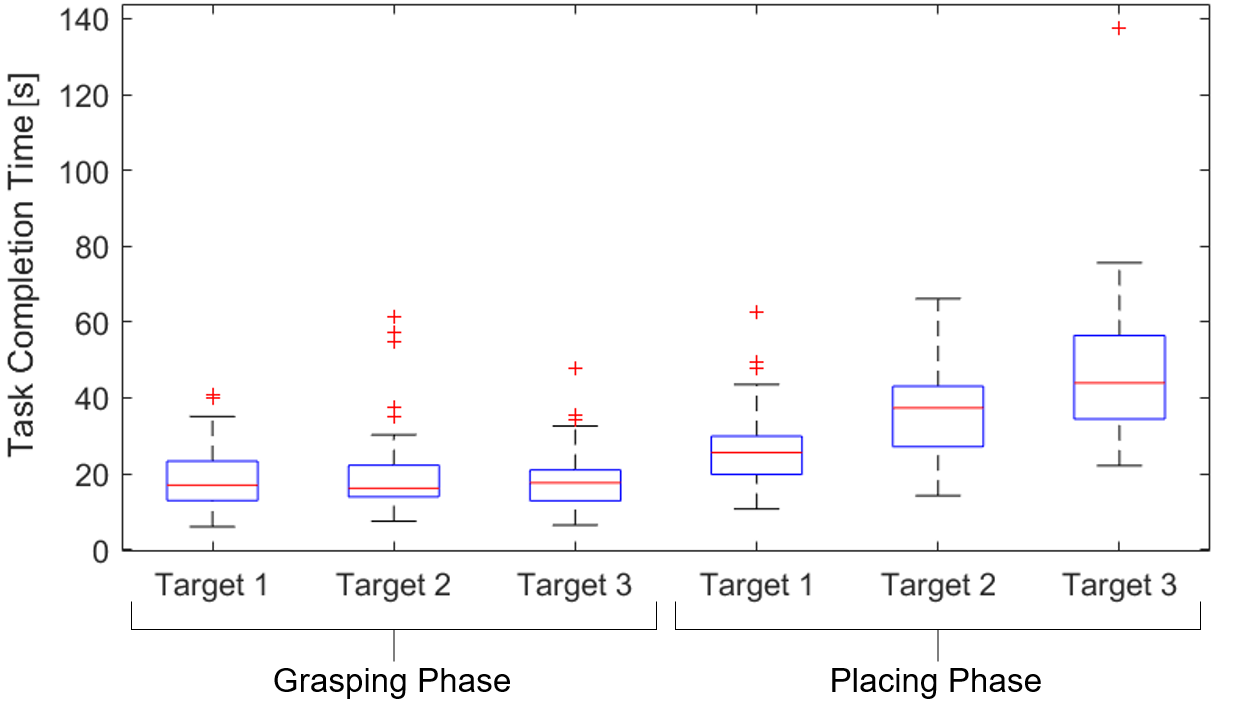}}\qquad
	\subfigure[\protect\url{}\label{fig:tct_steering}]%
	{\includegraphics[width=8cm]{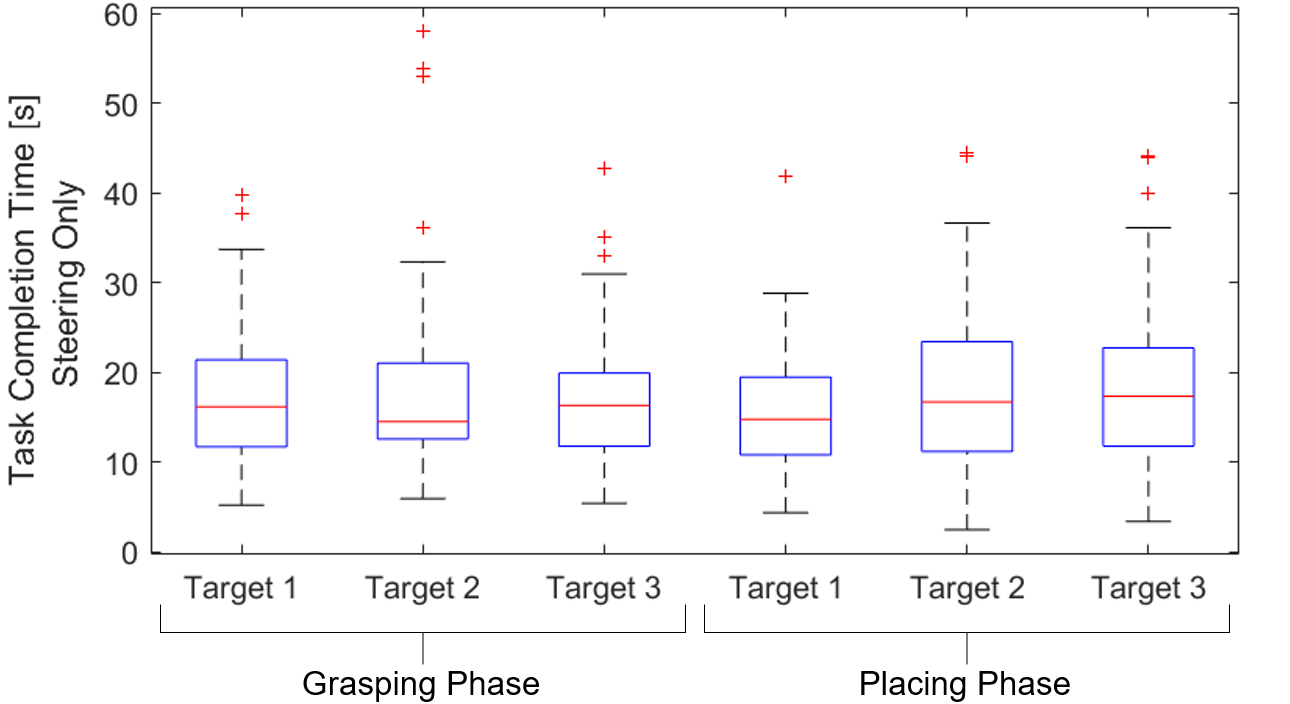}}
	\caption{Task Completion Time data for each target and each phase, including median, interquartile range with outliers, and max/min across all participants. In (a) is reported the overall time from start to end; in (b) is reported the time spent in steering, which is the time spent to tune the final position.}
	\label{fig:tct}
    \vspace{-0.55cm}
\end{figure}

Fig.~\ref{fig:errorHistogram} shows the histogram of the Target Placement Error over all trials, considering all the participants and all the targets: most of the trials resulted in errors lower than $2$~cm, and in particular, three of them presented a minimum of $0.2$~cm, showing that our system can achieve accurate performance in a manipulation task when operated by the Body Interface.
The histogram also shows the Failure Rate: only $5$ trials were discarded against $180$ successful performances, which translates to a success rate of $97\%$.

%Learning
Since both the Body Interface and the growing robot were new to the participants at the start of the experiment, we verified if the control was intuitive enough to learn during the short training period, or if additional learning took place over the course of the trials. We plotted how performance metrics for both phases of the experiment, Grasping and Placing, changed through the 15 trials (Fig.~\ref{fig:learning}). For Grasping, we measured performance using the time it took to grasp the block. Fig.~\ref{fig:learning_1} shows the Grasping results for all participants and the average across participants, showing that there was steady performance during the experiment, with only a very slight improvement in the average time to grasp. For Placing, we used Target Placement Error (TPE) as the measure of performance. Fig.~\ref{fig:learning_2} shows that the average placement error remained steady and low (below $2$~cm) throughout the experiment. Both results show that the interface was intuitive for users to operate and the majority of learning took place in the initial training block.

%Movements of the wrist in 3D
To understand the participants' performance of the task beyond these performance metrics, we examined how participants commanded the robot to reach all three targets. Fig.~\ref{fig:executionPath} shows three examples of the path performed by the participants while executing the task, one for each target. Each plot illustrates the best performance in terms of timing for the respective target, and shows the path within the workspace of the participant based on the Body Interface calibration (the values of the axes are expressed in millimeters). Furthermore, the deadband $[db_l, db_u]$ is also shown to show where on the path the growth and retraction commands were triggered.
These plots indicate that the strategy followed by the participants was mostly consistent from target to target, and followed the instructions provided during the training phase. During the Grasping phase, participants started the trial growing towards the block, tuned the position of the gripper by steering and then performed the grasp (black $\times$); subsequently, during the Placing phase, they retracted the robot to avoid any collision with the pillar, and then moved towards the target while steering and growing, ultimately tuning the position for the best placement.

We can focus on the results shown in Fig.~\ref{fig:executionPath} in two ways: in the breakdown in Task Completion Time, and in the location of block placement in the users' command space. As suggested by Fig.~\ref{fig:tct_total}, the Placing phase took more time than the Grasping one; this is true especially for Target 3 (see also Fig.~\ref{fig:movT3}), which was the furthest from the starting height of the cube and required more growing time. However, as shown by Fig.~\ref{fig:tct_steering}, if we do not consider the time spent during eversion (growth and retraction), there are no noticeable differences between Grasping and Placing. This indicates that steering was equally easy at all lengths when using the Body Interface. 

\begin{figure}[t!]
  \centering
	{\includegraphics[width=.95\linewidth]{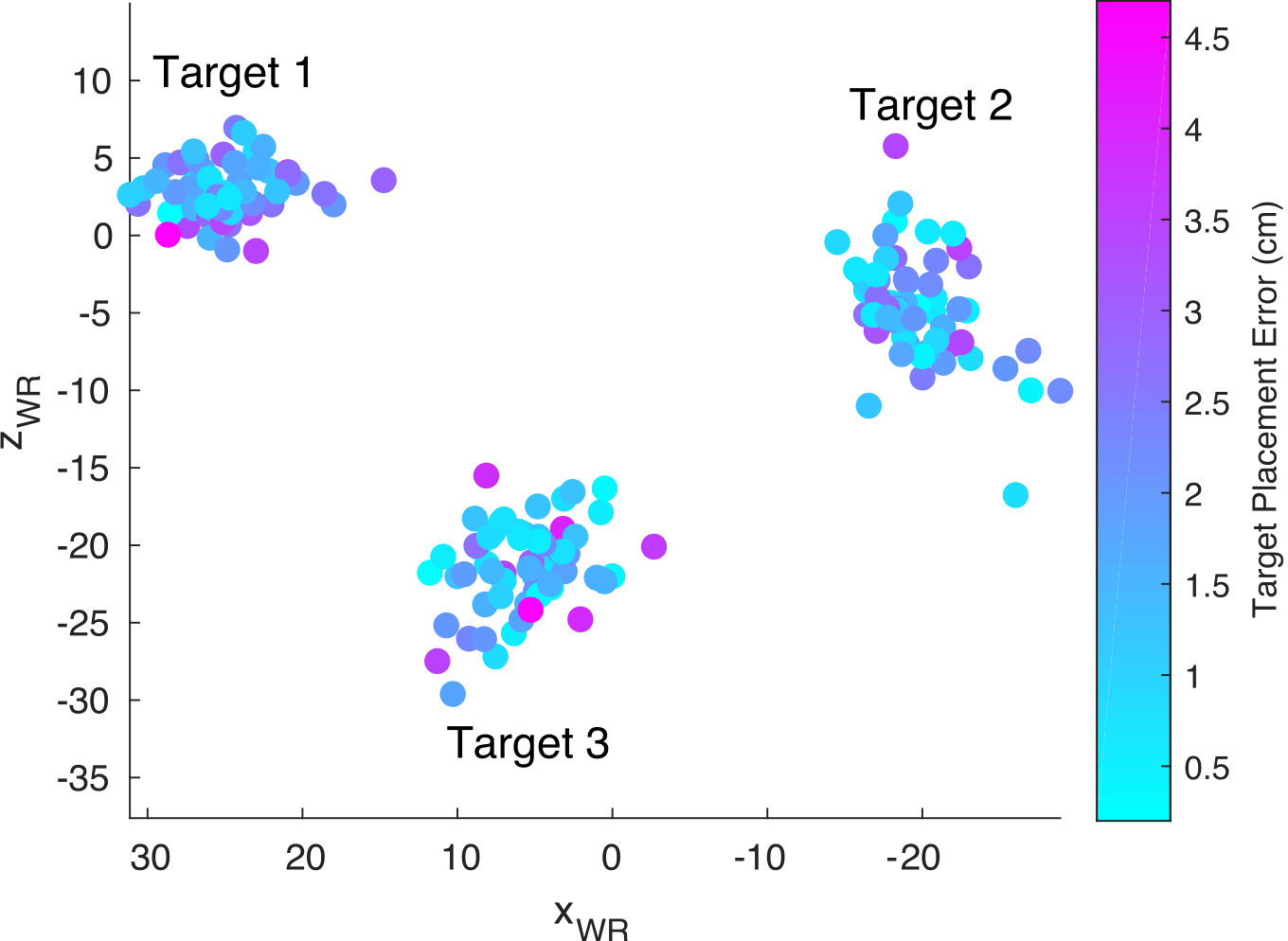}}
    \caption{Locations of placement command in users' calibrated wrist coordinates for the three targets. Color of the placement location indicates the Target Placement Error.}
    \label{fig:commandSpace}
    \vspace{-0.4cm}
\end{figure}

Looking at the locations in the command space where participants placed the block, we can see clear clusters indicating each of the three target locations (Fig.~\ref{fig:commandSpace}). The color of the dots indicate the Target Placement Error for that trial. We can see two interesting features in the data: some high error placements occurred close to the center of the clusters, and some low error placements occurred well outside. The high error dots can be explained by the quality of the grasp for that trial, since some grasps would cause the block to roll or move significantly after being released. On the other hand, even though participants were instructed to place the block on the target, the low error dots outside the clusters show where participants did not grow the robot as far and dropped the block from a height. This strategy required a larger steering command from the participant to reach the same location vertically over the target, putting them outside the cluster.

%Placement location

%NASA-TLX
Lastly, the results of the NASA-TLX showed an average workload value of $68\pm11\%$ among all the participants, which indicates that the task was challenging but not overly demanding. The participants indicated that the Grasping phase was slightly more difficult than the Placing: it was easy to hit the block with the gripper when trying to align the robot well, especially after overshooting the growth command. 

\section{CONCLUSION}
\label{sec:conclusion}

%summary
In this paper, we presented an intuitive interface to teleoperate a soft growing robot with arm gestures. 
We demonstrated that this interface can be used to perform a pick-and-place task by users with no previous training, and that those users can achieve placement errors below $2$~cm on average. 
This work shows a promising first step for creating interfaces that allow humans to control soft robots more intuitively and close the loop around the nonlinearities between joint and task space.

%future works 1 - soft gripper
In the future, we would like to improve the performance of the soft robot for teleoperated manipulation. Specifically, the participants consistently indicated that the Grasping phase was the hardest part of the task. We believe that the primary reason is the two-finger gripper design used, and the need to align it precisely to the block surface. A possible solution is to integrate a more compliant and adaptable gripper, like a four-fingered soft gripper \cite{ilievski2011soft}; such a device would ensure a more powerful and stable grasp, and remove the need to accurately position the gripper in the pronosupination degree of freedom. 

%future works 2 - flexible joystick and communication speed
Another important extension of this study would be to compare the Body Interface with previous proposed control interfaces, specifically the flexible joystick proposed by El-Hussieny et al. \cite{el2018development}. 
%To do so, the task should be modified appropriately to exploit the steering capability of the robot to assume different shapes, which is a strength point of the flexible joystick. 
%\todo{Furthermore, the Body Interface needs to be improved in terms of communication speed: the serial communication and the microcontroller (\textit{Teensy 3.6}) created a bottle neck that the joystick can easily overcome, running at $200$~Hz. However, the current experiment showed promising results even at a lower frequency of $66~Hz$, not a very noticeable difference in terms of human perception. NOTE $\rightarrow$ is this ok or way to risky to say? }

%future works 3 - shared autonomy
Finally, the last extension of the work will be to develop shared autonomy protocols to improve the interaction during teleoperation. The results of this work have shown that, although the Body Interface can achieve good performance in terms of accuracy and timing, there is still room for improvement. By allowing the robot to participate in the execution of the task, the role of the human operator will be simplified and the different strengths of the human and the robot can be exploited.  Different strategies that could be examined include: (i) haptic feedback through a holdable device \cite{walker2019holdable}, allowing the robot to provide guidance information to the operator and suggest the correct path to reach the targets; and (ii) artificial-intelligence algorithms mimicking the assist-as-needed paradigm used in robot-based rehabilitation \cite{stroppa2018improved}, where the robot will move autonomously towards the target only when the operator needs help to finalize the movement and only of a limited magnitude. 


%%%%%%%%%%%%%%%%%%%%%%%%%%%%%%%%%%%%%%%%%%%%%%%%%%%%%%%%%%%%%%%%%%%%%%%%%%%%%%%%

%%%%%%%%%%%%%%%%%%%%%%%%%%%%%%%%%%%%%%%%%%%%%%%%%%%%%%%%%%%%%%%%%%%%%%%%%%%%%%%%

%%%%%%%%%%%%%%%%%%%%%%%%%%%%%%%%%%%%%%%%%%%%%%%%%%%%%%%%%%%%%%%%%%%%%%%%%%%%%%%%
%\section*{APPENDIX}


%%%%%%%%%%%%%%%%%%%%%%%%%%%%%%%%%%%%%%%%%%%%%%%%%%%%%%%%%%%%%%%%%%%%%%%%%%%%%%%%

\bibliographystyle{IEEEtran}
\bibliography{references}

\end{document}